\pdfoutput=1

\documentclass[11pt,a4paper]{article}

\usepackage[acceptedWithA]{tacl2021v1}

\usepackage{times}
\usepackage{latexsym}
\usepackage[T1]{fontenc}
\usepackage{url}

\usepackage[final]{microtype}
\usepackage{booktabs}
\usepackage{enumitem}

\usepackage{datatool}
\DTLsetseparator{,}
\DTLloaddb{stats}{dataset-stats.csv}
\newcommand{\datavar}[1]{\DTLfetch{stats}{datavar}{#1}{value}}

\usepackage[disable]{todonotes}

\usepackage{xcolor}

\title{Dubbing in Practice: A Large Scale Study of Human\\ Localization With Insights for Automatic Dubbing}

\author{William Brannon\thanks{{ }{ }Work conducted during an internship at Amazon.} \\
  MIT Media Lab \\
  wbrannon@mit.edu
  \\\And
  Yogesh Virkar \\
  AWS AI Labs \\
  yvvirkar@amazon.com 
  \\\And
  Brian Thompson\thanks{{ }{ }Corresponding author.} \\
  AWS AI Labs \\
  brianjt@amazon.com
}

\date{}

\begin{document}
\begin{filecontents*}{dataset-stats.csv}
datavar,value
num_en_aligned,"204,734"
pct_en_aligned,87.37
num_es_aligned,"23,209"
pct_es_aligned,80.81
num_de_aligned,"26,099"
pct_de_aligned,89.35
num_gender_annotated,"23,304"
pct_gender_annotated,54.39
original_en_num_episodes,674
original_en_num_shows,54
original_en_num_speakers,"9,215"
original_en_num_hours,319.57
original_en_num_lines,"234,322"
original_es_num_lines,"28,720"
original_de_num_lines,"29,210"
post_filter_num_episodes,688
post_filter_num_shows,52
post_filter_num_hours,355.36
post_filter_num_lines,"201,246"
final_num_episodes,49
final_num_shows,11
final_num_hours,35.68
final_num_lines,"42,850"
final_pct_on_off_annotated,9.68
episodes_dropped_by_offset,3
words_with_alignment_problems_pct,12
\end{filecontents*}

\maketitle

\begin{abstract}

We investigate how humans perform the task of dubbing video content from one language into another, leveraging a novel corpus of \datavar{original_en_num_hours} hours of video from \datavar{original_en_num_shows} professionally produced titles.
This is the first such large-scale study we are aware of. 
The results challenge a number of assumptions commonly made in both qualitative literature on human dubbing and machine-learning literature on automatic dubbing, arguing for the importance of vocal naturalness and translation quality over commonly emphasized isometric (character length) and lip-sync constraints, and for a more qualified view of the importance of isochronic (timing) constraints. 
We also find substantial influence of the source-side audio on human dubs through channels other than the words of the translation, 
pointing to the need for research on ways to preserve speech characteristics, as well as semantic transfer such as emphasis/emotion, in automatic dubbing systems. 
\end{abstract}

\section{Introduction}\label{sec:intro}

Considerable attention has been paid to the dubbing of video content from one language into another, both in the literature of several disciplines and in the daily practice of the entertainment industry. 
One influential line of work, in the fields of film studies and audiovisual translation, studies human dubbing from a qualitative perspective \cite{chaumeAudiovisualTranslationDubbing2012, zabalbeascoaDubbingNonverbalDimension1997, zabalbeascoaNatureAudiovisualText2008, freddiAnalysingAudiovisualDialogue2009}, as a profession and semiotic activity. This literature has developed a rich body of theory on the nature of the human dubbing task, and the ways humans approach it, but has little engagement with large-scale data. More recently, machine-learning practitioners have taken up the task of building multimodal systems for automatic dubbing (e.g., \citet{sabooIntegrationDubbingConstraints2019,federicoSpeechtoSpeechTranslationAutomatic2020,tamIsochronyawareNeuralMachine2022}), but lack deep empirical or theoretical bases for how to organize their work.

What is missing from both literatures, and can help bridge the gap between them, is a large-scale study of human dubbing in practice: a data-driven examination of the way humans actually perform this task. Such an analysis can have benefits for both the qualitative study of human dubbing, by providing empirical evidence of how dubbing teams approach their work, and informing future machine-learning work on automatic dubbing systems. It is exactly this analysis we undertake in this work.

Human dubbing involves a sequence of human contributors each with control over a different aspect of the process \cite {chiaroIssuesQualityScreen2008, chiaroIssuesAudiovisualTranslation2008, matamalaTranslationsDubbingDynamic2010, chaumeAudiovisualTranslationDubbing2012}. The first step is an approximately literal translation of the original script, done by a dialogue translator. Next, a dialogue adaptor will modify this  translation into a plausible script meeting the requirements for dubbing such as isochrony, lip-sync, kinesic synchrony, and so on. Finally, the translated and modified script will go to a production team. Voice actors, with input from a dubbing director or supervisor, have been noted to often have freedom to improvise or make small changes to the dialogue as it is being recorded \cite{paolinelliTradurreDoppiaggioTrasposizione2009, chiaroIssuesQualityScreen2008, matamalaTranslationsDubbingDynamic2010}.

We, however, aim to understand human dubbing by studying not its process, but its product: a large set of actual dubbed dialogues from real TV shows, obtained from Amazon Studios. As compared to qualitative work or interviews with dubbers, this approach has the particular virtue of capturing tacit knowledge brought to bear in the human dubbing process but difficult to write down or explain.

We organize our investigation around one of the most fundamental insights from the qualitative literature, that of human dubbing (and subtitling, which we do not consider here) as "constrained translation" \cite{titfordSubtitlingConstrainedTranslation1982, mayoralConceptConstrainedTranslation1988}. A dub, after all, is not just a translation of the original content -- indeed it is not a purely textual product at all. As a translation, it should preserve the meaning of the original; as spoken language, it should sound natural; as an accompaniment to a video track, it should fit with the timing of actors' mouth movements, body language and the flow of the story \cite{chaumeDubbing2020}.

Simultaneously satisfying all of these constraints is very difficult, and in general may not be possible. We are accordingly interested in how human dubbers balance the competing interests of semantic fidelity, natural speech, timing constraints, and convincing lip-sync. Each can be traded off against the others, with varying effects on the audience's experience of the resulting product.

We operationalize this broad question as several more specific ones about the human dubbing process:\footnote{We do not consider other constraints or synchronies, like cultural fit with the target audience; though such constraints are important, they are too difficult to examine quantitatively.}
\begin{description}[noitemsep,topsep=0pt]
    \item[Isochrony] Do dubbers respect timing constraints imposed by the video and original audio?
    \item[Isometry] Do the original and dub texts have approximately the same number of characters?
    \item[Speech tempo] How much do voice actors vary their speaking rates, possibly compromising speech naturalness, to meet timing constraints?
    \item[Lip sync] How closely do the voice actors' words match visible mouth movements of the original actors?
    \item[Translation quality] How much will dubbers reduce translation accuracy (i.e. adequacy and fluency) to meet other constraints?
    \item[Source influence] Do source speech traits influence the target in ways not mediated by the words of the dub, indicating emotion transfer?
\end{description}

After exploring each of these questions, we provide insights on several research directions to address weaknesses we uncover in current automatic dubbing approaches.

\section{Related Work}
\label{sec:related}

\subsection{Qualitative}
\label{subsec:related-qualitative}
Modern qualitative research on human dubbing began with a seminal monograph by \citet{fodorFilmDubbingPhonetic1976}, himself a translator and writer of dubbed dialogue. He explored many of the constraints and methods which later literature elaborated.

Dubbing (both human and automatic) has subsequently come to be viewed as a type of \textit{constrained translation}, with more constraints than settings like comics, songs, or voice-over video content \cite{mayoralConceptConstrainedTranslation1988}. Most of the constraints stem from the need for a close match to the original video track.

In particular, dubs have \textit{isochronic} constraints: they should be about the same duration as the source, and should respect perceptible pauses within a speaker turn \cite{miggianiDialogueWritingDubbing2019}. Similarly, dubs benefit from complying with \textit{phonetic synchrony}\footnote{This is the term used in the literature, but synchrony of visemes \cite{fisherConfusionsVisuallyPerceived1968} would be a more accurate name, as lip sync bears on externally visible mouth movements.} or \textit{lip sync}: compatibility between the articulatory mouth movements required to produce the dub and the mouth movements, when visible, of the original actors \cite{fodorFilmDubbingPhonetic1976, miggianiDialogueWritingDubbing2019}.

Dubs also need to consider \textit{kinesic synchrony}: the plausibility of the dubbed dialogue in light of visible body movements of the original actors \cite{chaumeAudiovisualTranslationDubbing2012}. These three constraints -- isochronic, phonetic and kinesic -- are true 'synchronies' in modern usage as they relate to time. Kinesic synchrony is also an example of the broader category of semiotic or iconic constraints, or constraints "inherent to film language" \cite{chaumeDubbing2020}: the need for coherence between the language of the dub and the visual information of the film \cite{martiferriolCineIndependienteTraduccion2010}.

Dubs, of course, have non-temporal constraints as well. As cultural products, they should be readily intelligible to a member of the target linguistic and cultural community, with foreign references avoided or used for effect. (As \citet{chaumeDubbing2020} puts it, they must comply with "sociocultural constraints.") As speech, they should sound natural, as though originally recorded in the target language. Dubs which fail to meet this criterion are often considered examples of "dubbese" \cite{myersArtDubbing1973}. The peculiarities of "dubbese" have been studied extensively in a wide range of language pairs (see \citet{herbstDubbingDubbedText1997}, \citet{nencioniParlatoparlatoParlatoscrittoParlatorecitato1976}, \citet{pavesiAllocuzioneNelDoppiaggio1996}, \citet{freddiAnalysingAudiovisualDialogue2009}, and many others), especially as it may be 
specific to a national or linguistic community of dubbers \cite{chaumeDubbing2020}.

Turning to content, dubs have the same goal as any translation of preserving the semantic meaning of the source. However, some leeway is allowed; \citet{chaumeDubbing2020} provides two examples: (1) In a Spanish-to-English dub, an off-screen omelet may be turned into a pie, as the word for pie better adheres to lip-sync constraints. (2) In a Japanese-to-English dub, non-visible chopsticks might be changed to a fork to adhere to sociocultural constraints. Viewed through this lens, dubbing is a form of non-literal translation called "transcreation" \cite{zanottiTranslationTranscreationDubbing2014}. However, it is often desirable to keep such changes to a minimum to preserve fidelity to the source film \cite{martiferriolCineIndependienteTraduccion2010}.

Finally, other qualitative research has examined the social and textual nature of dubbing \cite{bosseauxDubbing2018, chaumeDubbing2020}. Scholars have investigated the role of power, ideology, identity, and similar considerations, in the production of dubs \cite{miggianiDialogueWritingDubbing2019, demarcoAudiovisualTranslationGender2012, santamariaFilmmakingCulturalReferents2016}.

\subsection{Automatic Dubbing}
\label{subsec:related-automatic}
Several works have explored the automatic generation of dubs, focusing on a variety of constraints.

One line of work has focused on integrating lip sync constraints into the dub generation process. \citet{taylorMouthFullWords2015} developed a method for automatic dubbing that matches the visemes of the original speech. \citet{sabooIntegrationDubbingConstraints2019} integrated lip-sync constraints into an encoder-decoder machine translation architecture. Taking a different approach, \citet{kimNeuralStylepreservingVisual2019} have explored adjusting mouth movements in the original video to match a dubbed audio track.

Other literature has examined "isometric" machine translation: producing a translation for use in automatic dubbing which has a similar length (in characters) to the input. It's argued that this property is "a proxy for the duration of its spoken realization" \cite{lakewMachineTranslationVerbosity2021}, 
and that similarity in character length makes TTS-generated speech sound more natural \cite{lakewISOMETRICMTNeural2022}. 
This approach has garnered interest from the community in the form of a shared task at IWSLT 2022 \cite{anastasopoulosFindingsIWSLT20222022}.

A third line of work has focused on controlling the speaking rate in automatic dubbing systems to achieve \textit{prosodic alignment}, or "synchronizing the translated transcript with the original utterances" \cite{federicoEvaluatingOptimizingProsodic2020}. \citet{oktemProsodicPhraseAlignment2019} focused only on the linguistic content matching between source-target phrases as a way to improve TTS, while  \citet{federicoSpeechtoSpeechTranslationAutomatic2020} focused on fluency. 
Their subsequent works \cite{federicoEvaluatingOptimizingProsodic2020, virkarImprovementsProsodicAlignment2021} further enhanced prosodic alignment by addition of features controlling for TTS speaking rate variation and linguistic content matching. 
Additionally, they introduced a time-boundary relaxation mechanism that can help to control speaking rate and speech fluency. 
\citet{virkarProsodicAlignmentOffscreen2022a} extended the time-boundary relaxation to further relax timing constraints for sentences that are off-screen.
\citet{tamIsochronyawareNeuralMachine2022} examined integrating pause constraints directly into MT. 
Finally, in contrast to the pipeline architecture used in most automatic dubbing works, \citet{huNeuralDubberDubbing2021} explored end-to-end dubbing.

\subsection{Empirical Studies}
\label{subsec:related-empirical}
In recent years, some studies have attempted to examine human dubbing through a quantitative lens, providing empirical information to inform theoretical debate. One line of work, such as \citet{sanchez-mompeanProsodyDubbedSpeech2020, sanchez-mompeanDubbingProsodyInterface2020}, has done detailed studies of prosody in human dubs, generally in a language-specific way. 
Other recent work has employed laboratory eye-tracking studies \cite{peregoEMPIRICALTAKEDUBBING2016} to gauge audience reaction. \citet{digiovanniChapterAreWe2019}, in particular, found that audiences may not be as sensitive to lip sync as traditionally believed. 
They report the existence of a "dubbing effect", in which audiences subconsciously avoid looking at the mouth movements of on-screen actors when dubbed speech fails to be lip synced. 

In the ML literature, recent work by \citet{karakantaTwoShadesDubbing2020} concluded that on-screen human dubs have significantly lower translation quality (i.e. translation adequacy and/or fluency) than human off-screen dubs, with the drop in quality attributed to the need to satisfy constraints (e.g., isochrony) not applicable or less applicable to offscreen dubs. They draw this conclusion -- on the HEROES corpus \cite{oktemBilingualProsodicDataset2018} -- by training a show-specific MT system and showing that it has lower performance (as measured by BLEU against the human dub) for on-screen than off-screen.

\section{Corpus Description \& Preprocessing}
\label{sec:data}
We begin with a dataset 
consisting of every TV show produced by Amazon Studios which was available on Prime Video at the end of 2021 for which we were able to locate a hand-curated transcript (for English shows) or dubbing script (for dubbed shows). 
These scripts are produced as part of the human dubbing process (see \autoref{sec:intro} for more details). 
This dataset contains \datavar{original_en_num_episodes} episodes of \datavar{original_en_num_shows} shows, comprising \datavar{original_en_num_hours} hours of content from \datavar{original_en_num_speakers} distinct speakers.
A detailed summary of this dataset is provided in \autoref{table:orig_data}.
Prime Video reports one or more genres per show - to provide more insight into the characteristics of this data, we report statistics for all genres for which we have at least 400 lines of manual on/off annotations (see \autoref{subsec:data-onscreen} for more details): Drama, Kids, Comedy, and Suspense. These subsets are used extensively in future sections to check the robustness of our conclusions. Note that these genre subsets have some overlap, due to the fact that some shows have more than one reported genre. 

All shows were originally recorded in English; we acquired both audio and video for the English originals and audio tracks for the professionally produced Spanish and German dubs where available. 
Much of our analysis relies on a subset of \datavar{final_num_hours} hours of content with both Spanish and German dubs. Our dataset also includes final transcripts from both the original and dubbing videos, which contain dialogue lines read by original or voice actors, with each line having a timestamp or "timecode" indicating its relative start time within the episode.

We perform extensive quality filtering prior to analysis. Data amounts for the entire corpus as well as each genre/language subset, at each stage of processing/filtering described below, are provided in \autoref{table:funnel}.

\begin{table*}
\begin{center}
\begin{tabular}{llrrrrr}
\toprule
Genre & Language &  Episodes &  Shows &  Speakers &   Duration (hrs) &   Dialogue Lines \\
\midrule
ALL                &   English  &       674 &     54 &      9,215 &  319.6 &  234,322 \\
                   &   German   &        72 &     13 &      2,498 &   43.2 &   29,210 \\
                   &   Spanish  &       197 &     18 &      7,384 &  118.7 &   28,720 \\
\midrule
Drama              &   English  &       264 &     23 &      4,737 &  161.5 &  115,549 \\
                   &   German   &        39 &      7 &      1,817 &   29.2 &   18,892 \\
                   &   Spanish  &       132 &     10 &      5,809 &   93.7 &   22,125 \\
\midrule
Kids               &   English  &       320 &     17 &      2,086 &  113.2 &   82,508 \\
                   &   German   &        32 &      5 &        654 &   13.7 &    9,972 \\
                   &   Spanish  &        60 &      7 &      1,483 &   23.0 &    6,224 \\
\midrule
Comedy             &   English  &       157 &     16 &      2,449 &   74.1 &   61,080 \\
                   &   German   &        23 &      4 &      1,023 &   14.8 &   13,146 \\
                   &   Spanish  &        58 &      6 &      2,197 &   30.4 &    7,942 \\
\midrule
Suspense           &   English  &        52 &      6 &      1,002 &   33.1 &   23,008 \\
                   &   German   &         8 &      2 &        336 &    5.7 &    3,338 \\
                   &   Spanish  &        19 &      3 &        902 &   12.9 &    3,985 \\
\bottomrule
\end{tabular}
\end{center}
\caption{Number of episodes (``Episodes,'' e.g. a 45min video), shows (``Shows'', e.g. a show might have 2 seasons each with 10 episodes), The number of speakers as estimated by the number of distinct characters in the given show (``Speakers''), and total run time for the show (``Duration''),
and the number of distinct dialogue lines (``Dialogue Lines'') for the show.
We report statistics for the entire corpus (``All'') as well as four genres (Drama, Kids, Comedy, and Suspense), 
in each of the 3 languages considered in this work (English, the source language, as well as German and Spanish dubs).
}\label{table:orig_data}
\end{table*}

\subsection{Segmentation and Forced Alignment}
\label{subsec:data-seg-align}
The first step of our data preparation pipeline uses script timecodes to segment audio tracks. As scripts do not include end times, each dialogue line is associated with the audio between its start time and the start time of the next line (or the end of the episode for the last line). Lines are roughly, but not exactly, the same as speaker turns: sometimes one line is only part of a speaker turn, and more rarely one line may include multiple speakers or crosstalk. The timecode-based segmentation process produces \datavar{original_en_num_lines} dialogue lines for English, \datavar{original_de_num_lines} for German, and \datavar{original_es_num_lines} for Spanish.

Next, we use the Montreal Forced Aligner \cite[MFA]{mcauliffeMontrealForcedAligner2017} to 
force align each dialogue line with its corresponding audio, producing a sequence of phones spoken in each word, along with start and end times for each phone.
MFA successfully aligns \datavar{pct_en_aligned}\% of English lines (\datavar{num_en_aligned}), \datavar{pct_de_aligned}\% of German lines (\datavar{num_de_aligned}) and \datavar{pct_es_aligned}\% of Spanish lines (\datavar{num_es_aligned}). 

Speaker fundamental frequency (i.e. F0 or, less formally, ``pitch'') is extracted using pyworld\footnote{\url{https://github.com/JeremyCCHsu/Python-Wrapper-for-World-Vocoder}} and linearly interpolated to fill in missing values, and energy is computed from Mel spectrograms of the speech signals. Both pitch and energy are averaged on a per-phone basis. 

\subsection{Filtering} 
\label{subsec:data-filtering}
There are several ways our data collection, segmentation, and alignment procedures might fail. We extensively filter the English side of the dataset to identify and remove erroneous dialogue lines.
Specifically, we filter out the following:

\textbf{Foreign-language text} We identify dialogue lines in the English originals whose text is not in English.
We use a language identification model for text\footnote{\url{https://huggingface.co/papluca/xlm-roberta-base-language-detection}}
and exclude anything with a low probability of being English,
as well as one entire show whose script text appeared not to be in English.

\textbf{Foreign-language audio} Similarly, we identify lines with non-English audio (from original non-English speech and errors in the corpus), using an audio language identification model trained on the VOXLINGUA107 corpus from the SpeechBrain toolkit \cite{ravanelli2021speechbrain, valkVOXLINGUA107DatasetSpoken2021}. We excluded an entire show whose supposedly English audio was actually German, several characters who spoke only in non-English languages, and any lines with low probability of being English.
	
\textbf{Multiple speakers or overlapping speech} Because overlapped speech is likely to confuse MFA, we ran overlapped speech detection \cite{bredinPyannoteAudioNeural2020, bredinEndToEndSpeakerSegmentation2021}, and excluded anything with a detected fraction of overlap higher than 30\%.

\textbf{Incorrect alignments} We performed ASR on each line's audio using an in-house tool 
and excluded dialogue lines with a) empty ASR output, b) an exact match to the gold text except for an inclusion at the front (these indicate segmentation errors), or c) a Levenshtein distance to the original greater than 80\% of the original length.

After filtering, we have \datavar{post_filter_num_lines} dialogue lines, from \datavar{post_filter_num_episodes} episodes and \datavar{post_filter_num_shows} shows, comprising \datavar{post_filter_num_hours} hours of source and target content. Manual inspection with Praat \cite{praat} suggests that post-filtering alignment quality is acceptably high. Most words and phones are correctly aligned, with only \datavar{words_with_alignment_problems_pct}\% of words in a hand-audited sample containing any phones with major problems. Errors most frequently occurred on foreign words and at silence boundaries, with word-initial or word-terminal phones incorrectly aligned into a preceding or following silence.

\subsection{Cross-lingual alignment}
\label{subsec:data-cross}
Given sets of force-aligned and filtered content in each language, we still need to align across languages to create a single corpus of parallel (English, dub) examples. 

\textbf{Offset finding} Many of the dubbed shows have episode-initial inserts, such as recaps of previous episodes or intro segments. Our dataset lacks target-side videos, and in lieu of manually identifying these segments from the audio tracks, we rely on cross-correlation of the aligned speech signals. For each (English, dub) pair of a given episode we sample at 100Hz a binary indicator of whether MFA has aligned a non-silence phone, and find the offset that maximizes the cross-correlation of these two signals. By inspection, these offsets work well and produce closely correlated patterns of silence between English and dubbed content. This process revealed \datavar{episodes_dropped_by_offset} episodes with quality issues, which were dropped from further analysis.

\textbf{Sentence alignment} Finally, we need to align groups of sentences occurring at approximately the same time in each (English, dub) episode pair. Note that because voice actors do not have to respect the exact distribution of silences in the original audio track, we have a many-to-many alignment problem: many stretches of speech in English may correspond to many, indeed potentially a different number, of stretches of speech in the dub. Accordingly, we align mostly according to content, using the Vecalign algorithm \cite{thompsonVecalignImprovedSentence2019, thompson-koehn-2020-exploiting} on multilingual LASER embeddings \cite{artetxeRobustSelflearningMethod2018} of the English and dubbed lines. 
We align contiguous stretches of speech. After alignment, we perform two final filtering steps to remove any spurious alignments, dropping sentence pairs where: either duration is exactly one frame, or the midpoint times of the source and target speech segments differ by more than 1s.

The final dataset of parallel cross-lingual alignments contains \datavar{final_num_lines} aligned dialogue line pairs, from \datavar{final_num_episodes} episodes and \datavar{final_num_shows} shows, comprising \datavar{final_num_hours} hours of source and target content.

\subsection{Gender Annotations}
\label{subsec:data-gender}
We extract information from the \textit{dramatis personae} lists in the original scripts on characters' genders. The scripts do not list all characters from whom we have speech, and differences in name formatting mean that some characters' gender information is lost. We are able to collect gender annotations for \datavar{num_gender_annotated} dialogue lines (\datavar{pct_gender_annotated}\% of the filtered corpus).

\subsection{On-Screen Annotations}
\label{subsec:data-onscreen}
We used annotations in the German dubbing scripts to identify on-screen (i.e., when the actor's mouth is visible) and off-screen (the actor's mouth is not visible) speech in the \datavar{final_num_episodes} episodes which had English, Spanish and German versions. Because these are the scripts actually used by dubbing professionals, they are not only human judgments of when characters' mouths are visible, but also directly influenced the actual human dubbing process. Only approximately \datavar{final_pct_on_off_annotated}\% of aligned pairs have on-screen/off-screen annotations.

Because much of our analysis rests on comparing onscreen and offscreen dialogue lines, we would also like to test for systematic differences in what type of content is onscreen or offscreen. In particular, we look for statistically significant differences in the duration of onscreen and offscreen lines, and (to measure the complexity of speech) the average perplexity of the GPT-2 language model \cite{radfordLanguageModelsAre2019} on each set of lines. Reassuringly, neither is significantly different: an independent-samples t-test fails to reject the null hypothesis that onscreen and offscreen examples have the same source-side mean duration ($p = 0.106$), and bootstrapping the average GPT-2 perplexity fails to reject the null hypothesis that it is the same between onscreen and offscreen ($p = 0.08$). It is possible that dubbing professionals themselves skip adding on/off annotations in cases (like narration) when it would be obvious from the text itself whether it is onscreen or offscreen.

\begin{table}
\centering
\begin{tabular}{p{1.3cm}rrrr}
\toprule
Subset &  Orig &  Filter &  Align &  On/Off  \\
\midrule
ALL                  &            292,252 &               201,246 &          42,850 &       3,617 \\
\midrule
Drama                &            156,566 &               115,159 &          27,845 &        3,097 \\
Kids                 &             98,704 &                72,938 &          14,351 &        446 \\
Comedy               &             82,168 &                54,525 &          21,034 &        2,278 \\
Suspense             &             30,331 &                20,789 &           5,046 &       608 \\
\midrule
German                 &           29,210 &                25,739 &            22,892  &   1,926 \\
Spanish                &           28,720 &                23,196 &            19,958  &   1,691 \\
\bottomrule
\end{tabular}
\caption{Total number of dialogue lines, for various stages of filtering, for all of the data (``ALL''), genre subsets (Drama, Kids, Comedy, and Suspense), and both target language subsets (German and Spanish).
The ``Orig'' column gives the number of lines before any filtering (see \autoref{sec:data}).
The ``Filter'' column gives the number of lines after quality filtering described in sections 
\autoref{subsec:data-seg-align}
and
\autoref{subsec:data-filtering}.
The ``Align'' column gives the number of lines after cross-lingual alignment described in 
\autoref{subsec:data-cross}.
Finally, the ``On/Off'' column gives the number of lines which have manual on-screen / off-screen annotations (see \autoref{subsec:data-onscreen}). 
}\label{table:funnel}
\end{table}

\subsection{Data Release Considerations}
\label{subsec:data-release}
Unfortunately, content licensing restrictions prevent us from releasing our data. We believe this will be the case for any similar high-quality, large corpus: professionally written, acted, produced, and dubbed shows are proprietary for commercial reasons.

Note that a few prior works \cite{pavesiChapter14Dubbing2009,oktemBilingualProsodicDataset2018} have released human dubbing datasets;
however, these datasets are much smaller than the dataset considered in this work and the legality of these datasets relies on a very permissive interpretation of ``fair use'' which may not be acceptable at some organizations.

\section{Analysis}
\label{sec:analysis}

\subsection{Isochrony}
\label{subsec:analysis-isochrony}

Perhaps dubs' most obvious constraint is isochronic: dubbed speech should line up in time with the original speech.
This constraint is especially binding when the character's mouth is visible ("onscreen"), but may apply for other reasons even when it is not ("offscreen"): examples include cuts or transitions in the video, surrounding onscreen speech, and the need to align with actors' body movements. Many qualitative works have considered isochronic constraints (e.g., \citet{chaumeAudiovisualTranslationDubbing2012, miggianiDialogueWritingDubbing2019, fodorFilmDubbingPhonetic1976}), and automatic dubbing work has explored integrating them, usually with a proxy for isochrony such as length in syllables \cite{sabooIntegrationDubbingConstraints2019, oktemProsodicPhraseAlignment2019} or in characters \cite{federicoSpeechtoSpeechTranslationAutomatic2020, lakewMachineTranslationVerbosity2021,lakewISOMETRICMTNeural2022,tamIsochronyawareNeuralMachine2022}. We are accordingly interested in exploring how much human dubbers respect this constraint.

First, we simply compare the durations of aligned dialogue line pairs. Source duration ought to be a strong predictor of dub duration, which indeed it is: the correlation\footnote{Correlations are Pearson unless otherwise noted.} between the two is quite high at $r = 0.877$. But duration does not consider the actual start and stop times of lines, and may simply reflect the need to convey the same amount of information in source and target. 

As a further check, we look at the \textit{overlap fraction} of speech time: the amount of time in each dialogue line when \textit{both} the original (source language) actor and the dubbing voice actor (target language) are speaking (i.e. the intersection), divided by the amount of time when \textit{either} is speaking (i.e. the union). 
A value of $1.0$ indicates perfect time alignment, while $0.0$ indicates the source and target speech occur at entirely different times. The mean overlap fraction in our corpus is 0.658
and the median is 0.731 -- in 4.3\% of lines, overlap is exactly 0, pulling down the mean. Thus, while human dubbed speech mostly co-occurs with source speech, 
isochronic constraints are also frequently violated by human dubbers.

We observe that on-screen dubs are more isochronic than off-screen, 
but to a surprisingly small degree. The average offscreen dialogue line has overlap fraction 0.662, vs 0.684 on-screen -- an increase of only 3.3\%.
Excluding animated shows where characters' (animated) mouth movements may be less constraining, this gap rises slightly: offscreen overlap of 0.656, vs. 0.690 onscreen, for an off-to-on increase of 5.2\%. Both differences, while small in magnitude, are statistically significant at the $\alpha = 0.01$ level under independent-samples t-tests (overall: $t = -2.93$, $p = 0.003$; live-action: $t = -5.35$, $p = 8.7 \times 10^{-8}$).
For individual genre subsets, we find the offscreen to onscreen gap is significant for Drama ($t = -2.95$, $p = 0.003$) and Comedy ($t = -3.70$, $p = 0.0002$) but not for Kids and Suspense.
The increase is not significant at the $\alpha=.01$ level for either language (German: $t = -2.28$, $p = 0.02$, Spanish: $t = -1.57$, $p = 0.12$).
\footnote{For significance tests in this work, unless otherwise noted we tests for the entire corpus for which the test is valid, as well as for subsets of the valid corpus corresponding to each target language (German and Spanish) and the genres listed in \autoref{table:orig_data} (Drama, Kids, Comedy, and Suspense).}

The small gap in on- vs off-screen isochrony may be partially explained by our on/off screen annotations: the dubbing professionals are likely only annotating sections where on- and off-screen dialogues are mixed, and the on-screen constraints may be constraining preceding/successive off-screen lines.

\subsection{Isometry}
\label{subsec:analysis-isometry}

Past work \cite{lakewISOMETRICMTNeural2022, anastasopoulosFindingsIWSLT20222022} has examined similarity of text length (measured in characters) as a way to constrain translation for automatic dubbing, especially a requirement that the target translation be within $\pm 10\%$ of the source character length. This practice is called "isometric machine translation" \cite{lakewISOMETRICMTNeural2022}, and we refer to the length constraint as `isometry.' This literature uses isometry mainly as a proxy for similarity of duration and for isochrony, though it may also help avoid large variations in TTS output rates \cite{lakewISOMETRICMTNeural2022}. We aim to test these assumptions: how good a proxy is isometry for isochrony in human dubs, and how much do human dubbers preserve character length?

We examined the text length (measured in characters) of aligned (source, human dub) dialogue line pairs, and especially the percentage change in character length from source to dub. Character lengths on both sides included punctuation and spaces (except at the start or end of a dialogue line). To measure how well character length similarity proxies for similarity of duration, we compared the ratio of target to source length to the ratio of target to source duration. We examine here only known onscreen lines, as these are subject to the greatest pressure to be isochronic; results are very similar if using all lines.\footnote{We exclude from analysis dialogue lines where either source or target had an aligned duration less than 0.2s; by inspection, most of these lines are segmentation errors.} 

We find first that isometry is a weak-to-moderate proxy for isochrony. The ratio of human dub to English character lengths has a correlation of only $r = 0.279$ ($r^2 = 0.078$) with the time overlap fraction of source and target, though it is somewhat more correlated with the ratio of target to source durations ($r = 0.620$, $r^2 = 0.385$).

\begin{figure}[ht]
    \centering
    \includegraphics[scale=0.52]{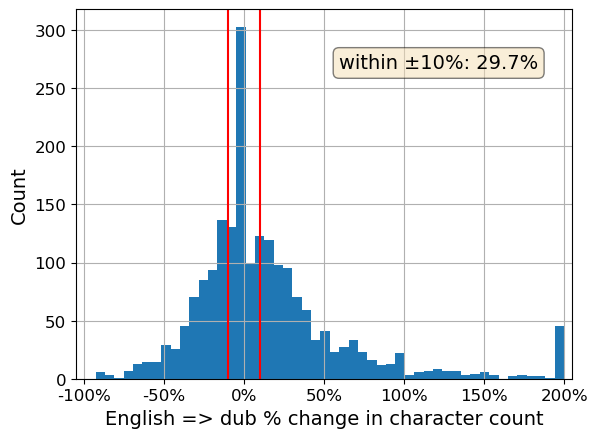}
    \caption{Percentage change in character count from English source to human dub among onscreen lines. The plot is clipped at a 200\% increase (and the character count can't decrease by more than 100\%). Vertical red lines indicate a $\pm 10\%$ change.}
    \label{fig:character_count_change}
\end{figure}

Our results on character length similarity in human dubs, meanwhile, are summarized in \autoref{fig:character_count_change}. Overall, there are large changes in character length from source to human dub. Most sentence pairs differ in length by more than prior work's 10\% threshold. The absolute percentage change in character length is significantly different from 0 under a one-sample t-test ($t = 20.3$, $p = \textrm{3.9e-83}$). 
These changes are significant for both languages and all genre subsets with $p < \textrm{1e-38}$.
Character lengths are more similar for longer sentences (and the distribution of character count change is smoother), but nearly 60\% of pairs in which the source sentence is at least 50 characters long differ in length by more than 10\%.

We observe that human dubs are largely non-isometric in both Spanish and German,
with neither language differing from the English source lines by more than 10\% in less than 69\% of cases. The length differences are, however, distributed differently. German skews toward longer dub lines, with 53\% of all lines longer than the matched English lines by >=10\%, and 16\% shorter by >=10\%; Spanish displays a smaller skew in the opposite direction, with 30\% at least 10\% longer and 42\% at least 10\% shorter.

\subsection{Speaking Rate}
\label{subsec:analysis-speaking-rate}

Previous literature has paid considerable attention to the naturalness of human dubbed speech \cite{sanchez-mompeanProsodyDubbedSpeech2020}. A frequent, though not universal, conclusion is that dubs sound "artificial and contrived" \cite{chaumeDubbing2020}, for reasons ranging from strange intonations to "anglicisms" inspired by the source language \cite{frescoNaturalnessSpanishDubbing2009}. 
From another angle, the isometric MT literature argues that TTS models, which are less flexible than humans in varying speaking rate, may require isometric input to produce natural sounding isochronic output \cite{lakewISOMETRICMTNeural2022}. Because naturalness is a broad topic, and in general may require human evaluation, we focus on examining speaking rates. We're particularly interested in whether dubbing voice actors are willing to vary their speaking rates, and perhaps compromise naturalness, in order to meet other constraints, like isochrony.

We examined both the dub speaking rate\footnote{In this work, we calculate speaking rate as the average number of words spoken per second in each dialogue line, including pause time, following \citet{laverPrinciplesPhonetics1994}.}
and the ratio of human dub duration to source duration as functions of the number of words in source and target dialogue lines.\footnote{As in \autoref{subsec:analysis-isometry}, we excluded lines with either source or target duration less than 0.2s. Results are robust to thresholds as low as 0.06s. We also exclude 4 dialogue lines, which appear to be alignment errors, in which the target-to-source duration ratio was more than 20.} As the dub-to-source ratio of word lengths increases, in other words, what happens to dub speaking rate and the duration ratio?

Perhaps counterintuitively, it seems that the duration ratio is much more closely related to relative length of content than the dub speaking rate. Simple linear regression of each outcome variable on the word length ratio indicates a correlation of 0.523 between word length ratio and duration ratio ($r^2 = 0.273$), while human dub speaking rate has a correlation of only 0.163 with duration ratio ($r^2 = 0.027$).

As an additional check, we examined the variance of speaking rate (at the dialogue-line level) on source and human dub. If the dubbing voice actor is varying speaking rate to meet timing constraints, we would expect more variability in the dubbed speech than the source speech. 
We do not, however, observe this: the standard deviation of dubbing voice actor speaking rate is lower for both Spanish (1.25 w/s, vs 1.47 w/s on the source side) and German (1.26 w/s, vs 1.46 w/s on the source side). In both cases we can reject the null hypothesis that the standard deviation of speaking rate is higher for dub than source via a percentile-bootstrap test (Spanish: $p < 10^{-10}$; German: $p < 10^{-10}$). Likewise, we can reject the null hypothesis for all genre subsets considered with $p < 10^{-10}$.

When forced to pick one or the other, human dubbers appear more willing to break timing constraints than vary speaking rate.

\subsection{Lip Sync}
\label{subsec:analysis-lip-sync}

Both qualitative \cite{chaumeAudiovisualTranslationDubbing2012, fodorFilmDubbingPhonetic1976, miggianiDialogueWritingDubbing2019} and technical work \cite{taylorMouthFullWords2015, huNeuralDubberDubbing2021, kimNeuralStylepreservingVisual2019} have considered "lip sync" constraints in human and automatic dubbing, respectively. 
The idea is that dubbed audio should match the (visible) mouth movements of the original actors. Failing to do so may be jarring to the audience and reduce the quality of the dub. Some recent empirical studies, however, have found that this constraint may not be as binding as previously assumed \cite{peregoEMPIRICALTAKEDUBBING2016}. Accordingly, we ask here whether human dubbers produce speech which matches the mouth movements of the original actors. 

Rather than relying on the video tracks, we use the notion of a "viseme", or visual phoneme \cite{fisherConfusionsVisuallyPerceived1968}, to capture alignment between source and human dub mouth movements. Phones in the same viseme are produced with similar articulatory movements of the lips and tongue, and look visually similar. We use viseme tables\footnote{\url{https://docs.aws.amazon.com/polly/latest/dg/ref-phoneme-tables-shell.html}} to map each MFA-aligned phone to its corresponding viseme. The glottal stop and four guttural German sounds made without moving the lips are dropped. To measure cooccurrence, we sample the viseme active on both source and dub sides
and compute the viseme-viseme cooccurrence matrix. We normalize the matrix so that the observed frequency of each viseme pair is a fraction of the frequency expected if source and target visemes were independent, but with the observed marginal distributions. 

Over all the data (onscreen, offscreen, and unannotated), the average within-viseme cooccurrence rate as a fraction of the rate under independence is 1.575, with an average across-viseme rate of 0.981. For  onscreen the average within-viseme cooccurrence rate is 1.613, while for offscreen it is 1.463. We believe both on- and off-screen rates are above 1.0 due to the presence of names and cognates, where the phones may be (nearly) the same for the source and dub.
The difference is statistically significant at the $\alpha = 0.05$ level under a percentile-bootstrap test ($p = 0.017$) for the entire corpus,
as well as for both languages and all genre subsets.

Moreover, we see similar patterns by language: The amount of excess cooccurrence is 40.8\% for German (1.638 to 1.898)\footnote{The increase in excess cooccurrence is (0.898 - 0.638) / 0.638 = 40.8\%.} and 37.8\% for Spanish (1.439 to 1.605).
But though the effect is significant, it is not large in absolute terms: even in onscreen speech, only about 12.4\% of speech time has the same viseme
on the source and target sides. This suggests that human dubbers do sometimes lip-sync their output, but it is a fairly soft constraint.

Note that this analysis is sensitive to small errors in the exact start and stop times of aligned phones, and to inconsistencies across languages in the phone boundaries used to train aligners. Our results are thus likely to be a lower bound on how closely human dubbers observe lip sync constraints.

\subsection{Translation Quality}
\label{subsec:analysis-quality}

As previously discussed (see \autoref{sec:intro}), the human dubbing process is complicated, with the translation modified throughout the process to satisfy isochrony, lip-sync, and other constraints. Thus an obvious question to ask is how faithful the resulting translation actually is to the source material.\footnote{We use the term \emph{translation quality} here to refer to translation adequacy and fluency. It does not refer to the overall quality of the human dubbing output, which may intentionally decrease translation adequacy and/or fluency to meet other constraints.}
Reducing translation quality may, after all, make it easier to satisfy other constraints: for example, changing the meaning in the target language may better fit the original English mouth movements than a more correct translation. 

To get at this question, we rely again on the onscreen/offscreen annotations. Isochronic constraints should be more binding onscreen, and we've shown above that the (inherently onscreen) lip sync constraints are at least partly followed. If translation quality is sacrificed to meet these other goals, we should see lower-quality translations onscreen than offscreen.

In contrast to \citet{karakantaTwoShadesDubbing2020}, we use a more straightforward approach of simply measuring the quality of the human dubs using automatic MT metrics. Since we do not have access to the original, pre-adaptation, human translation, we rely on reference-free metrics. In particular, we measure performance for each (source, human dub) pair with comet-qe \cite{reiCOMETNeuralFramework2020} and prism-src \cite{thompsonAutomaticMachineTranslation2020, thompson-post-2020-paraphrase}. Despite the lack of references, both have been shown to have better correlation with human judgements of MT quality than BLEU \cite{papineni-etal-2002-bleu}, which requires a reference, in many cases \cite{thompsonAutomaticMachineTranslation2020, freitagResultsWMT21Metrics2021}.

\begin{figure}
    \centering
    \includegraphics[scale=0.6]{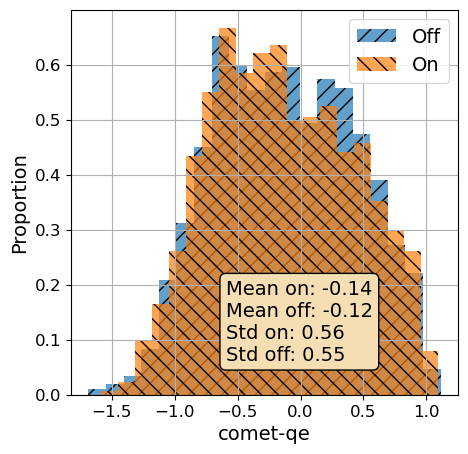}
    
    \caption{Comet-qe scores, including means and standard deviations, broken out by onscreen/offscreen status. Note that these scores represent only data points with onscreen/offscreen annotations. No meaningful onscreen/offscreen differences are observed in comet-qe or prism-src (not shown for space) scores.}
    \label{fig:comet-prism-semantic-change}
\end{figure}

The results are summarized in \autoref{fig:comet-prism-semantic-change}. We find no substantial differences between onscreen and offscreen speech for either metric, with the on/off difference in means being less than 1/10th of a standard deviation for both comet and prism. Neither average comet scores ($t = 0.936$, $p = 0.349$) nor average prism scores ($t = 1.51$, $p = 0.131$) were significantly different onscreen from offscreen under a two-sided independent-samples t-test. 
Likewise, for both metrics, we did not find statistically significant differences for either language or any genre at the $\alpha = .05$ level.
Though not depicted in \autoref{fig:comet-prism-semantic-change}, the results are similar when broken out by language: neither Spanish nor German dubs show any meaningful worsening of translation quality when onscreen. Human dubbers do not, in other words, appear to be sacrificing translation quality to hit other constraints.

\subsection{Non-Text Transfer}
\label{subsec:analysis-srctgt}

Finally, we explore whether the human dub audio depends on the source audio in ways not mediated by the text of the dub translation.

We first look at source influence on three aspects of dubbing speech actor (target language) audio: speaking rate, pitch, and energy. 
For pitch and energy, we compute both the mean and the standard deviation per dialogue line, relying on higher standard deviation, and thus greater range, of pitch and energy as a crude indicator of emotion \cite{frickCommunicatingEmotionRole1985}.
(We drop from analysis of standard deviations any line with only one phone on the source or target side.)
We also use the gender annotations extracted in \autoref{subsec:data-gender} to control for the effect of gender on dubbing voice actor's vocal pitch. As in \autoref{subsec:analysis-isometry}, we exclude dialogue lines where either source or target has aligned duration less than 0.2s.

Overall, we find that source audio properties explain a substantial fraction of target variance. Source speaking rate correlates with target speaking rate ($r = 0.439$, $r^2 = 0.193$), and the correlation is stronger the longer the dialogue lines. For lines with source and target both at least 1s long, the correlation is $r = 0.584$ ($r^2 = 0.342$). Line-level mean pitch is even more strongly related, with $r = 0.792, r^2 = 0.628$, though standard deviation of pitch is less so ($r = 0.429, r^2 = 0.184$). Both the mean ($r = 0.381, r^2 = 0.145$) and standard deviation ($r = 0.366, r^2 = 0.134$) of energy also display some linear relationship between source and target, though even more weakly.

By fitting sets of linear models predicting dubbing voice actor speaking rate, mean pitch and standard deviation of pitch, first as a function only of indicator / dummy variables for speakers, and second adding in the line-level property on the source side, we show this relationship is not simply a speaker-level effect: see \autoref{tab:audio_property_prediction}. While speaker identity is generally a good predictor of target audio characteristics, dialogue line-level variables also increase predictive power. This line-level information is more useful for speaking rate than pitch, but its increase in predictive power is significant for both. Additionally, we find the gender of the source character is only a weak predictor of line-level mean pitch, with an indicator variable for male having only about $r^2 = 0.058$ in predicting the dub-side mean pitch.

\begin{table}[ht]
    \centering
    \begin{tabular}{lrrr}
        \toprule
        Property & C(Speaker) & +Source & $\Delta$ \\
        \midrule
        Spk rate, 0.2s+ & 0.116 & \hspace{3mm}0.239 & +\textbf{0.122} \\
        Spk rate, 1s+   & 0.191 & 0.390 & +\textbf{0.198} \\
        Pitch mean      & 0.675 & 0.733 & +\textbf{0.057} \\
        Pitch std.      & 0.284 & 0.316 & +\textbf{0.032} \\
        Energy mean     & 0.184 & 0.268 & +\textbf{0.084} \\
        Energy std.     & 0.210 & 0.275 & +\textbf{0.065} \\
        \bottomrule
    \end{tabular}
    \caption{$r^2$ values for linear models predicting various properties of target audio (dubbing voice actor) from source audio (original actor). The first column reflects models containing only indicator or dummy variables for the speaker, while models in the second column add the line-level property for the source side. All increases in explained variance are significant at the $\alpha = 10^{-6}$ level by F-test. This finding also holds for both language and all genre subsets (not shown).}
    \label{tab:audio_property_prediction}
\end{table}

Altogether, these results suggest that there is quite a bit of both speaker-level and line-level influence for future machine learning work to consider.

\subsubsection{Semantic Transfer}
Finally, as a more stringent check, we also conduct word alignment via FastAlign \cite{dyerSimpleFastEffective2013} between each English dialogue line and its human dub. The alignment process produces (source, target) pairs of semantically similar words. If dubbing voice actors are imitating properties of the source speech in their own speech, we might expect to find that speaking rate, pitch and energy at the word level are more closely correlated within aligned word pairs than in other word pairs within the same dialogue line. We look in particular at the number of phones per second 
as well as the word-level mean and standard deviation of both pitch and energy.

\begin{figure}[ht]
    \centering
    \includegraphics[scale=0.37]{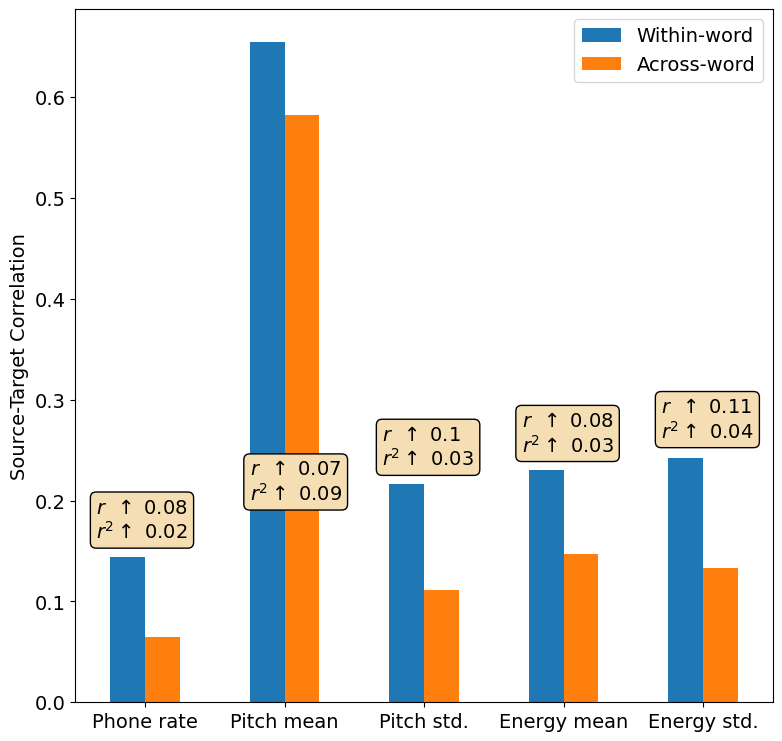}
    \caption{Pearson 
    correlations of various audio properties between source and target (dubbing voice actors) within aligned word pairs (``Within-word'') and within unaligned word pairs (``Across-word'') in the same dialogue line. All properties show greater correlation within aligned word pairs than across them.}
    \label{fig:word_alignment}
\end{figure}

All of these properties are, in fact, more closely correlated within aligned word pairs than in other word pairs, as shown in \autoref{fig:word_alignment}. The amount of increase from across-pair to within-pair ranges from 0.08 to 0.11, with all six increases significant at the $\alpha = 10^{-6}$ level by a test based on Fisher's z-transform for correlation coefficients. 
The increases are also significant for each language and genre subset at the $\alpha = 10^{-6}$ level.

As an even more stringent check, we first normalize the word-level variables, subtracting their line-level means and dividing by their line-level standard deviations. Doing so guards against the possibility that patterns at the line level, such as the amount of attention human dubbers pay to different types of line, influence these results.
This analysis, shown in \autoref{tab:word_alignment_norm}, confirms the findings of the unnormalized version. As expected, little to no relationship is visible between unaligned pairs of words, while aligned pairs are weakly, but positively, correlated across several metrics. All of these differences are also significant at the $\alpha = 10^{-6}$ level by the same test as above, using Fisher's z-transform. 
The increases are also significant for each language and genre subset at the $\alpha = 10^{-6}$ level.

\begin{table}[ht]
    \centering
    \begin{tabular}{lrrr}
        \toprule
        Property &  Within &  Across &  $\Delta$ \\
        \midrule
        Phone Rate  &   0.117 &  -0.009 & +\textbf{0.125} \\
        Pitch Mean  &   0.112 &  -0.009 & +\textbf{0.120} \\
        Pitch Std   &   0.090 &  -0.007 & +\textbf{0.097} \\
        Energy Mean &   0.092 &  -0.007 & +\textbf{0.099} \\
        Energy Std  &   0.108 &  -0.009 & +\textbf{0.116} \\
        \bottomrule
    \end{tabular}
    \caption{Pearson correlations of various audio properties between source and target (dubbing voice actor) within aligned word pairs ("within") and within unaligned word pairs ("across") in the same dialogue line. Word-level variables have first been normalized at the line level before being correlated, subtracting their line-level means and dividing by their line-level standard deviations.}
    \label{tab:word_alignment_norm}
\end{table}

These patterns clearly indicate that human dubbers are imitating properties of the source audio at a very granular (and \textit{semantic}) level. We interpret these results as evidence of emotion and/or emphasis transfer from source to target.

\section{Insights for Automatic Dubbing}
Our analysis of the human dubbing process points to several directions that should (and perhaps should not) be pursued in automatic dubbing, which we summarize below. 

\textbf{Translation quality} and \textbf{speech naturalness} appear to be paramount. The input to the dubbing process mostly consist of dialogue with challenging issues for automatic translation systems, like ambiguous speaker gender, ambiguous addressee gender and number, and formality between characters. Speaker gender and number issues are especially critical since the audience can often both hear and see the speakers and addressees. We note a stark lack of literature on automatic translation of dialogues, compared to common domains in literature like news. Likewise, naturalness for TTS systems is challenging enough under normal circumstances, but TV shows often include yelling, crying, whispering, etc, making the problem even harder. While research does exist in this space, we suspect there is much room for improvement.

We find strong evidence for several levels of \textbf{non-textual transfer} of source audio properties into human dubs: speaker characteristics, dialogue line-level effects, and emotion/emphasis transfer when considering \textit{semantic} alignments at the word level. This points to a glaring issue with pipeline approaches employed by the vast majority of automatic dubbing literature: Without a mechanism to encode emotion/emphasis, individual vocal profiles and other traits of the source speech, we expect them to be nearly impossible to replicate in synthetic target speech.

The high rates of \textbf{isochrony} that we observe in human dubs support the need for continued research on isochronic MT, especially given the observed unwillingness of human dubbers to vary their speaking rate,
which shows that automatic dubbing systems should not simply vary speaking rates to achieve isochronic constraints. 
However, our findings do not support the use of \textbf{isometric MT}.
Our work refutes the claim that isometry is a good proxy for isochrony, and it also casts doubts on the claim that isochrony is more necessary with TTS than with human voice actors because TTS is less able to vary speaking rates (i.e. we find that human dubbers are not varying speaking rate to meet isochronic constraints, and thus automatic dubbing systems should likely not either). The authors suspect that directly optimizing isochrony (as opposed to isometry) is likely a better approach for automatic dubbing. 

Finally, the low rates of \textbf{lip sync} that we observe (and the very small if still statistically significant difference between on- and off-screen rates) in human dubs suggest that research on automatic lip-sync can be marginally useful, at best, for automatic dubbing. Efforts like \citet{kimNeuralStylepreservingVisual2019}, however, which edit mouth movements in the video, may be an exception.

\section{Future Work}

This work focused on two language pairs: English-German and English-Spanish. In future work, we hope to analyze 
more distant language pairs (e.g. English-Chinese or English-Arabic), as well as non-English source material.

Our analysis has shown that isometry is a poor proxy for isochrony in human dubs, yet several prior works have claimed that isometric MT benefits automatic dubbing. In future work, we hope to perform analysis to understand this discrepancy.

The scope of this work necessitated automatic metrics. However, in future work, we hope to verify some of these findings (e.g. translation quality of on- vs off-screen) using human annotators.

Finally, the aggregate analysis in this work is necessary to provide high-level insights for automatic dubbing. However,
it likely also hides interesting variations across different individual translators, adaptors, dubbers, dubbing studios, etc. We hope to better explore these dimensions in future work. 

\section{Conclusion}

We present the first large-scale quantitative study 
of how humans perform the task of dubbing video content from one language into another.
Our results challenge a number of popular assumptions in both qualitative and machine learning literature: Human dubbers display less respect for isochrony and especially lip sync than is suggested by qualitative literature,
while being surprisingly unwilling to vary speaking rates or sacrifice translation quality to hit other constraints. 
Our analysis provides insights on research directions to address weaknesses in current automatic dubbing approaches.  

\section*{Acknowledgements}
This effort would not have been possible without Joel Chengottusseriyil, Robert Enyedi, Bradley
Gordon, Natawut Monaikul, Prashanth Rajagopal,
and Shuai Tang, who prepared the initial data used
in this work. We also thank Marcello Federico,
Prashant Mathur, Surafel Lakew, Reza Madad, and the anonymous TACL reviewers for helpful discussions and feedback.

\clearpage  
\bibliographystyle{acl_natbib}
\bibliography{
    references/references,
    references/anthology,
    references/brian
}

\end{document}